\documentclass[sigconf]{acmart}
\AtBeginDocument{%
  \providecommand\BibTeX{{%
    \normalfont B\kern-0.5em{\scshape i\kern-0.25em b}\kern-0.8em\TeX}}}
\AtBeginDocument{%
	\providecommand\BibTeX{{%
			Bib\TeX}}}

\acmSubmissionID{123-A56-BU3}

\usepackage{multirow}
\usepackage{cleveref}
\usepackage{tabularx}
\usepackage{color}
\usepackage{xcolor}
\usepackage{booktabs}

\usepackage[vlined,boxed,commentsnumbered,linesnumbered,ruled]{algorithm2e}
\usepackage{amsmath}

\usepackage{amssymb}

\usepackage[utf8]{inputenc}
\usepackage[T1]{fontenc}

\usepackage{tabularx}
\usepackage{adjustbox}
\usepackage{multirow}

\usepackage{colortbl}
\usepackage{listings}
\definecolor{green}{RGB}{113,165,55}
\definecolor{blue}{RGB}{1,158,213}
\definecolor{red}{RGB}{220,10,10}
\definecolor{LightGreen}{HTML}{d4edda}  
\definecolor{LightBlue}{HTML}{d1ecf1}   
\Crefname{table}{Table}{Tables}
\Crefname{figure}{Figure}{Figures}
\Crefname{equation}{Equation}{Equations}
\Crefname{paragraph}{Paragraph}{Paragraphs}

\begin{document}

\title{HMPE:HeatMap Embedding for Efficient Transformer-Based Small Object Detection}


 \author{Yangchen Zeng}
 \authornote{Both authors contributed equally to this research.}
 \affiliation{%
	   \country{China}
	   \city{Wuxi}
	 }
 \email{zengyangchen@foxmail.com}



\begin{abstract}

Current Transformer-based methods for small object detection continue emerging, yet they have still exhibited significant shortcomings. 
This paper introduces HeatMap Position Embedding (HMPE), a novel Transformer Optimization technique that enhances object detection performance by dynamically integrating positional encoding with semantic detection information through heatmap-guided adaptive learning.
We also innovatively visualize the HMPE method, offering clear visualization of embedded information for parameter fine-tuning.
We then create Multi-Scale ObjectBox-Heatmap Fusion Encoder (MOHFE) and HeatMap Induced High-Quality Queries for Decoder (HIDQ) modules. These are designed for the encoder and decoder, respectively, to generate high-quality queries and reduce background noise queries.
Using both heatmap embedding and Linear-Snake Conv(LSConv) feature engineering, we enhance the embedding of massively diverse small object categories and reduced the decoder multihead layers, thereby accelerating both inference and training.
In the generalization experiments, our approach outperforme the baseline mAP by 1.9\% on the small object dataset (NWPU VHR-10) and by 1.2\%\ on the general dataset (PASCAL VOC). By employing HMPE-enhanced embedding, we are able to reduce the number of decoder layers from eight to a minimum of three, significantly decreasing both inference and training costs.

\end{abstract}




\maketitle

\section{Introduction}
\label{sec:intro}
\begin{figure}[H] 
	\centering
	\includegraphics[width=1\columnwidth]{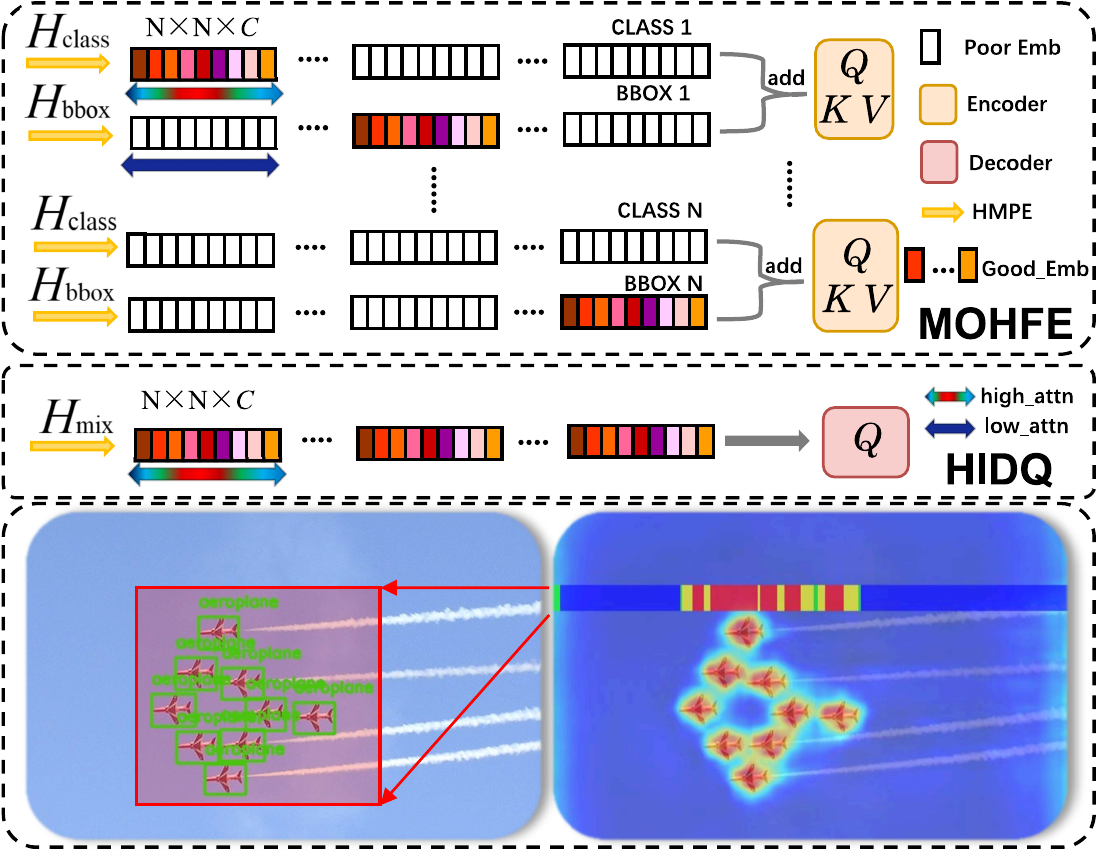}
	\caption{Heatmap embedding visualized with heatbar.The heatmap embedding shows "\textcolor{red}{hot}" middle and "\textcolor{blue}{cold}" ends, corresponding to "\textcolor{red}{good}" and "\textcolor{blue}{poor}" embeddings.}
	\label{fig:vision of HMPE}
\end{figure}
Small object detection has been a challenging yet fundamental task in computer vision for the last decade. It spans various vertical domains, including drone aerial images (UVAs), remote sensing images (RSI), satellite detection, infrared detection (IF), crater detection (CD), high-resolution wide (HRW) imagery, and more. Small object detection faces numerous distinct challenges, such as densely arranged targets, tiny objects, and high-noise background ratios, among others.

Compared to CNN-based detectors, Transformer-based detection has garnered more attention in the field of small object detection due to its higher performance ceiling.
In recent years, although Transformer-based models have made significant breakthroughs in the field of small object detection, existing methods still need improvement in the following aspects:

(1) Insufficient exploitation of position, class, and bounding box information during query generation in Transformer-based detection methods leads to suboptimal model performance when train with limited datasets.They fundamentally lack the exploration of intrinsic correlations between positional embedding and detection semantics.
(2) Poor detection quality stems from the lack of high-quality queries and failure to suppress low-score queries, jointly compromising both bounding box localization accuracy and category generation. 
(3) Decoder redundancy induces slow inference speeds. As multi-head attention complexity scales proportionally with model depth and input token length, the lack of effective information embedding necessitates excessive decoder heads, dramatically increasing computational costs for both training and inference.
(4) Detection of intricate small targets remains challenging due to the absence of dedicated feature engineering mechanisms for extracting discriminative characteristics from massively diverse small object categories.

To solve the above problems, we propose an innovative framework called  \textbf{HeatMap Embedding (HMPE)}, which dynamically aligns position embedding with object detection semantics through heatmap-guided adaptive learning.
By integrating heatmap augmentation into image embedding, we effectively incorporate rich semantic detection information through the heatmap, deeply unifying positional, class, and bounding box data. This significantly improves the detection head's precision in regressing geometric and semantic attributes while reducing the influence of background noise. 
It enables rapid convergence in network training and enhances the quality of network inference.
We also innovatively visualize the HMPE method, offering clear visualization of embedded information for parameter fine-tuning.

Utilizing HeatMap Embedding, we further propose two innovative structures to separately enhance the encoder-decoder architecture.
We introduce the \textbf{MOHFE} (Multi-Scale ObjectBox-Heatmap Fusion Encoder) and \textbf{HIDQ} (HeatMap Induced High-Quality Queries for Decoder) modules, leveraging HMPE to enhance the embedding information in both the encoder and decoder, respectively.
In the encoding phase, the MOHFE module innovatively integrates class and bbox semantics into the embedding, achieving multi-scale fusion through heatmap embeddings at various scales.
In the decoding phase, the designed HIDQ transforms mixed heatmap features into high-quality queries, enhancing the decoder by reducing redundant multi-head layers, thereby accelerating inference and improving training quality.
Then,to address the challenges posed by extreme feature sparsity in complex small targets and general detection tasks, we designed the \textbf{LSConv}(Linear-Snake Conv) module to capture these sparse features and embedded the extracted features into the network through HMPE.
In summary, the main contributions of this work are as follows:
\begin{itemize}
	\item We innovatively designe heatmap embedding and its visualization, enabling an observable insight into the embedded information, and utilized it to fine-tune the parameters.
    \item Leveraging heatmap embedding, we develope MOHFE and HIDQ specifically for the encoder and decoder, respectively, to generate high-quality queries.
    \item Using heatmap embedding and LSConv feature engineering, we reduce decoder layers, and speed up inference and training.
\end{itemize}

\section{Related Work}

\vspace{2mm}\noindent\textbf{CNN-based methods on small detection.}
With the rapid development of artificial intelligence and machine learning, traditional object detection methods have also achieved significant progress. Based on CNNs(Convolutional Neural Networks)\cite{he2016deep}, which excel at extracting representative features, tasks such as small object detection in remote sensing imagery have been successfully addressed. Traditional CNN-based object detection methods can be divided into two categories. The first is one-stage object detection, led by networks such as SSD\cite{liu2016ssd}, YOLO series\cite{redmon2018yolov3,zyc2022behavior}, and Retina-Net. The second is two-stage object detection based on region proposals, led by networks such as  Mask-RCNN\cite{he2017mask} and Faster R-CNN\cite{ren2015faster}.

Zhang et al. \cite{zhang2023superyolo} proposed the use of CNNs for solving the task of remote sensing image scene classification, where the model is capable of extracting deep features from remote sensing scene images. The PCNet \cite{cao2023pcnet} compares feature information differences between images to achieve scene classification. Chen et al.\cite{chen2025yolo} proposed a novel strategy that significantly enhances the multi-scale feature representation of real-time detectors.
\begin{figure*}[]
	\centering
	\includegraphics[width=1\textwidth]{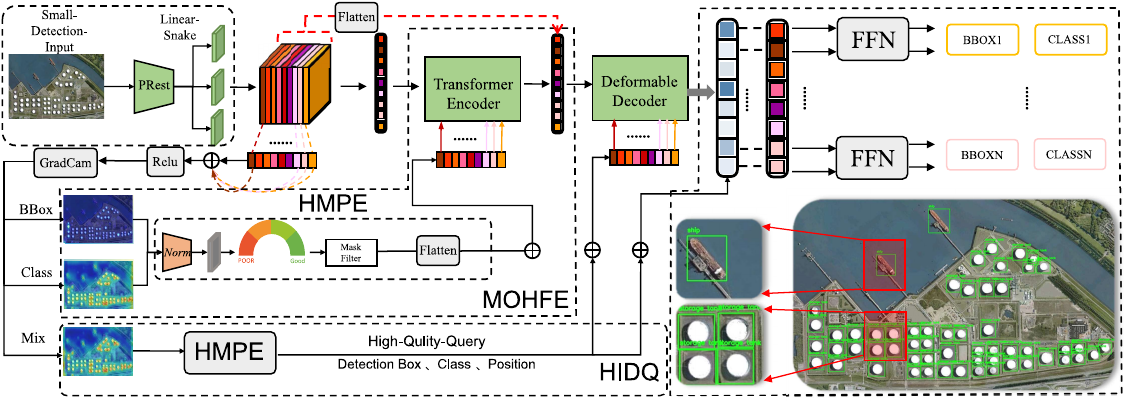}
	\caption{The pipeline of the proposed method HeatMap Embedding.It consists of the HIDQ, MOHFE module, and Linear-Snake.The HMPE module applies norm, followed by upsampling, and employs a Mask Filter to produce high-quality queries while filtering out low-quality background noise.}
	\label{fig:ppl-crop}
\end{figure*}

However, these methods generate a large number of candidate bounding boxes, typically requiring Non-Maximum Suppression (NMS)\cite{hosang2017learning} to filter them. NMS itself is a greedy algorithm that cannot guarantee a globally optimal solution, potentially leading to missed detections or false positives. Furthermore, the threshold for NMS needs to be manually set, often based on empirical experience and experimental outcomes, making it hard to generalize across different datasets and tasks,thereby extending both the training and inference time of the models.

\vspace{2mm}\noindent\textbf{Transformer-based methods on small detection.}
Transformers have achieved success in Natural Language Processing (NLP), and their potential in visual tasks has garnered significant attention. Compared to CNN-based methods, Transformer-based detection methods do not generate a large number of redundant bounding boxes, eliminating the need for subsequent Non-Maximum Suppression processing, which greatly reduces the likelihood of missing objects. Traditional Transformer-based object detection methods can be divided into two categories. The first is DETR architecture, led by networks such as DETR\cite{carion2020end}, RT-DETR\cite{zhao2024detrs}, ,cond-DETR\cite{meng2021conditional},Dino\cite{zhang2022dino} and Def-DETR\cite{zhu2020deformable}. These methods adopt an encoder-decoder architecture with CNN as the feature extractor, where CNN handles feature extraction, the encoder is responsible for feature fusion, and the decoder manages the regression of detection tasks. The second group encompasses Vision Transformer (ViT)\cite{rao2021dynamicvit}, which processes images in patches through a Transformer architecture to predict detection outcomes. This approach employs a Transformer-only framework for the task of image classification.

To address the issue of class imbalance, Yang et al. \cite{ma2024coarse} combined focal loss with Transformer and proposed a weakly supervised localization method based on CAMs(Class Activation Maps), allowing the model to better address the identification of regions containing prohibited objects.Li et al.\cite{li2024sparseformer} proposed a Sparse Vision Transformer, which significantly enhances the accuracy and speed of object detection in high-resolution wide-angle images by selectively focusing on sparsely distributed target windows and combining global and local attention mechanisms.

However, the computational cost of ViT for processing high-resolution images is impractical. Several approaches have been attempted to reduce the cost of ViT models, including window-based attention mechanisms \cite{liu2021swin}, downsampling~\cite{wang2021pyramid, wu2021cvt} in self-attention, and low-rank projection attention~\cite{xiong2021nystromformer}. Other works apply sparsity strategies to patches [36], self-attention heads [33], and Transformer blocks [33] in image classification.
These methods often experience a decline in precision or an increase in training costs when detecting small target objects.

\section{Proposed Method}

\subsection{HMPE:HeatMap Embedding And Visualization}
\label{subsec: HeatMap Embedding （HMPE ）and Visualization}

While DETR-based object detection methods exhibit remarkable advantages in global context modeling, they fundamentally lack the exploration of intrinsic correlations between positional embedding and detection semantics.
The absence of embedded information coupling—manifested as static positional embeddings and constrained synergy between semantic features—fundamentally limits further improvements in detection performance. As illustrated in the method pipeline figure~\ref{fig:ppl-crop} , this paper proposes the\textbf{ HMPE (HeatMap Embedding)} algorithm and implements HeatMap Embedding Visualization, which generates high-quality positional embedding integrated with detection semantics. 
This advancement significantly enhances query quality in encoder-decoder structure and improves small target detection accuracy.

Conventional approaches rely on uniform 2D Sinusoidal Positional Encoding schemes a technique that critically neglects multi-scale object variations and fails to decouple spatial dependencies between semantic and positional  features. Such limitations cascade into three deficiencies: loss of high-frequency positional information crucial for small targets, mutual interference between category classification and bounding box regression tasks, and inadequate high-dimensional semantic embeddings to distinguish densely distributed small objects in cluttered environments. These shortcomings collectively undermine detection robustness, particularly in scenarios requiring precise localization of miniature or overlapping instances.

To address this limitation, we propose HeatMap Embedding (HMPE) , a novel mechanism that generates high semantic density heatmap matrices through gradient-weighted class activation mapping , dynamically aligning positional embedding with semantic-critical regions in multi-scale detection tasks.

As illustrated in the pipeline framework Figure~\ref{fig:ppl-crop}.
\paragraph{Step 1} employs a convolutional neural network to extract input image feature tensors $\mathbf{A} \in \mathbb{R}^{K \times H \times W}$, where $A_k(i, j)$ denotes the activation intensity of the $k$-th filter at spatial coordinate $(i, j)$. 

For a detection class $c$, the \textbf{gradient weighting coefficients} $\alpha_{ijk}^{c}$ are derived by computing second- and third-order partial derivatives of the classification head's output confidence score $y^c$ with respect to the feature tensor $\mathbf{A}$:
\begin{equation}
	\alpha_{ijk}^{c} = \frac{\partial^2 y^c}{\partial A_k(i,j)^2} + \frac{\partial^3 y^c}{\partial A_k(i,j)^3}
\end{equation}

This formulation quantifies the nonlinear sensitivity of category confidence to spatial activation variations.
The global channel-wise importance weight $\beta_{k}^{c}$ for channel $k$ is computed via spatial aggregation incorporating ReLU-based gradient filtering:

\begin{equation}
	\beta_{k}^{c} = \sum_{i=1}^{H} \sum_{j=1}^{W} \alpha_{ijk}^{c} \cdot \text{ReLU}\left( \frac{\partial y^c}{\partial A_k(i,j)} \right)
\end{equation}

Here, $\text{ReLU}$ suppresses negative gradients to retain only activations positively correlated with class $c$, while $\alpha_{ijk}^{c}$ hierarchically amplifies regions where activation magnitudes exhibit higher-order nonlinear contributions to class discrimination.

\paragraph{Step 2}
Generate $H_{\text{class}}$ based on the channel-wise global importance weights $\beta_{k}^{c}$:

\begin{equation}
	H_{\text{class}}(i,j) = \text{ReLU}\left( \sum_{k=1}^K \beta_k^c \cdot A_k(i,j) \right)
\end{equation}

Analogous to $H_{\text{class}}$, replace the confidence score $y^c$ with the bounding box regression loss $L_{\text{reg}}$, and obtain activation weights through Huber loss gradient backpropagation to generate the detection-box heatmap $H_{\text{bbox}}$.
Generate the hybrid heatmap $H_{\text{mixed}}$ by dynamically adjusting the contribution ratio of categorical and geometric information via the gating mechanism $\lambda$:

\begin{equation}
	H_{\text{mixed}} = \lambda \cdot H_{\text{class}} + (1 - \lambda) \cdot H_{\text{bbox}}
\end{equation}

Obtain three upsampled masks with the same dimensions as the standard feature maps.

\paragraph{Step 3}
Construct a dynamic masking mechanism (\textbf{MASK Filter}) to decouple cold and hot regions in the heatmap, enabling heatmap-guided position embedding, as shown in Equation~\eqref{eqa:mask}. The mask matrix multiplies element-wise with standard positional encoding to suppress invalid positional noise in background regions while preserving multi-scale geometric features in heat regions, thereby enhancing semantic integration of class, position, and bounding box information in encoder-decoder architectures.

As shown in Figure \ref{fig:vision of HMPE}, the \textbf{heatmap embedding visualization} demonstrates background suppression through the gradient threshold of the mask filter $\mathit{Mask}$, which creates anisotropic attention patterns in hot/cold regions. Introducing ReLU gradient filtering during mask filter generation establishes a semantic screening mechanism:
\begin{itemize}
	\item For \textbf{Heat Area (Good Embeddings)} ($H_{\text{map}} > \tau$): retain full sinusoidal positional encoding. Executes precise geometric positional matching, visually depicted as heatmap focal points.
	\item For \textbf{Cold Area (Poor Embeddings)} ($H_{\text{map}} \leq \tau$): binarize positional embeddings via gradient heat-value filtering (ReLU suppresses invalid background noise). Degrades to global semantics, visually depicted as background.
\end{itemize}
Evaluation metrics demonstrate that the ReLU mask filter endows tokens with two characteristics: \textbf{Position branch} enhances semantically prominent global features, and \textbf{Detection branch} sharpens geometry-sensitive local variations.

This section introduces ~\textbf{Heatmap Embedding (HMPE)}, an innovative framework that dynamically aligns position embedding with object detection semantics through heatmap-guided adaptive learning. By guiding position embedding optimization through heatmap, it effectively enhances the collaborative performance of classification and localization in object detection.
The method significantly improves geometric detail perception in complex scenes and small object detection while reducing background interference. It provides a practical solution for decoupling semantic and positional features in visual tasks, thereby enhancing small object detection performance.

\begin{figure}[t] 
	\centering
	\includegraphics[width=0.95\columnwidth]{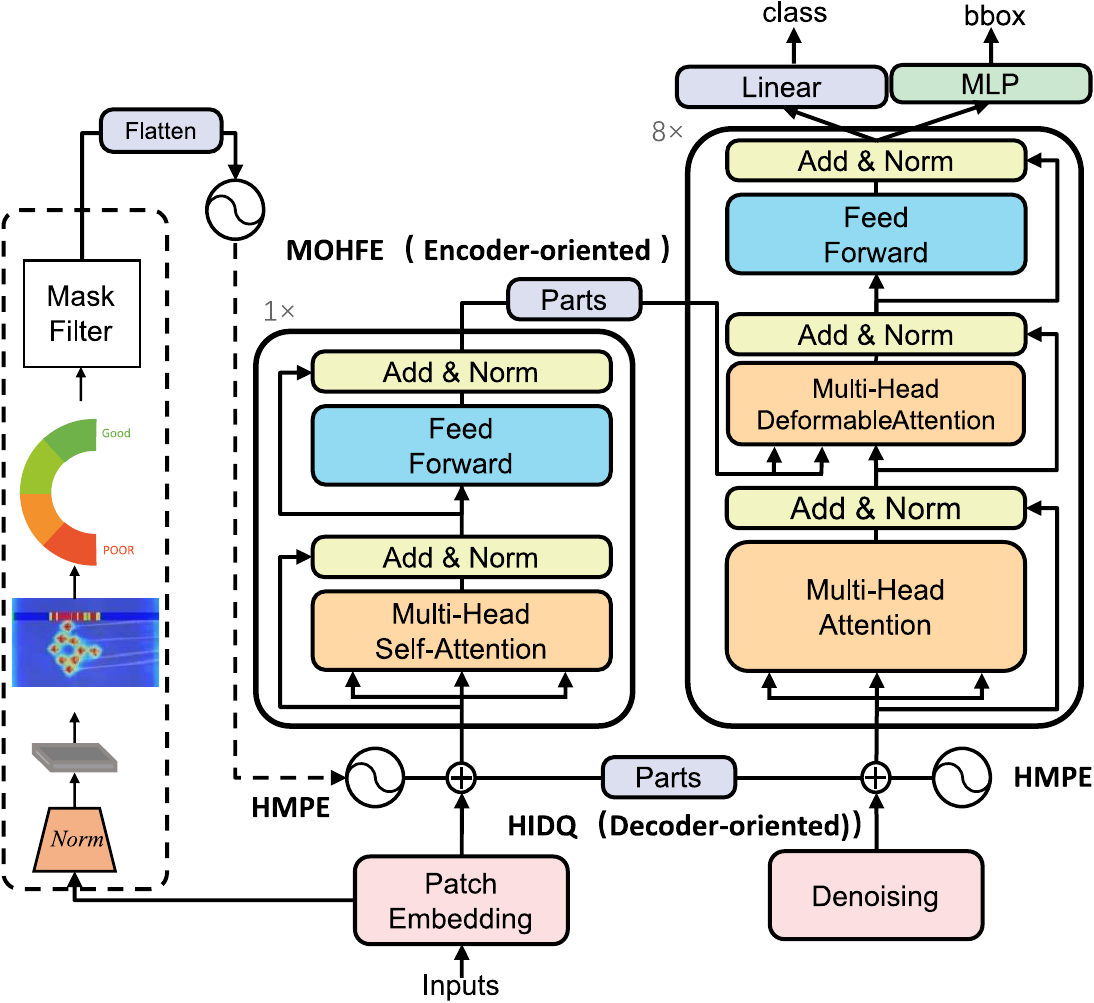}
	\caption{Transformer Optimization with HMPE.}
	\label{fig:all-crop}
\end{figure}

\subsection{HMPE to High-Quality-Query}
\label{subsec:HMPE to High-Quality-Query}

The HMPE method processes raw feature maps through multi-branch operations to generate three specialized heatmaps:category-predominant heatmap ($H_{\text{class}}$), bounding box-guided heatmap ($H_{\text{bbox}}$), and hybrid-guided heatmap ($H_{\text{mixed}}$). By leveraging the threshold property of the Mask filter, it filters Poor embeddings response regions, constructs a binarized mask matrix, and integrates high-quality queries generated from positional encoding. We propose the \textbf{MOHFE} and \textbf{HIDQ} modules to enhance the correlation between positional encoding and detection semantics in the encoder and decoder, respectively.

\subsubsection{MOHFE:Multi-Scale ObjectBox-Heatmap Fusion Encoder}

\label{subsubsec:MOHFE}

\textbf{MOHFE (Encoder-oriented)}:Its core innovation lies in establishing a conditional coupling mechanism between heatmap spectra and positional encoding, forging strong correlations between positional encoding and semantic features while enabling multi-scale feature fusion to expand receptive fields.

At the encoder stage, the MOHFE module applies independent linear projections to the category heatmap $H_{\text{class}}$ and bounding box heatmap $H_{\text{bbox}}$, concatenates the processed features $E_{\text{class}} \| E_{\text{bbox}}$, and generates Query/Key/Value matrices for multi-head attention:
\begin{equation}
	\begin{aligned}
		Q_{\text{enc}} &= W_Q [E_{\text{class}} \| E_{\text{bbox}}], \\
		K_{\text{enc}} &= W_K [E_{\text{class}} \| E_{\text{bbox}}], \\
		V_{\text{enc}} &= W_V [E_{\text{class}} \| E_{\text{bbox}}].
	\end{aligned}
\end{equation}

This process facilitates implicit high-level decoupling between semantic features and position features.

The mask matrix $\mathit{Mask}(i,j)$ zeros out positional embeddings for poor embeddings,
\begin{equation}
	\text{Mask}(i,j) = 
	\begin{cases} 
		0, & (i,j) \in \{ H_{i,j}^{\text{grad}} = 0 \} \\
		1, & \text{else}
	\end{cases}
 \label{eqa:mask} 
\end{equation}

 via:

\begin{equation}
	\text{PE}(i,j,d) = \text{Mask}(i,j) \odot \left[ \sin\left( \frac{i}{\tau_d} \right) + \cos\left( \frac{j}{\tau_d} \right) \right]
\end{equation}

as shown in the above equation, dynamically suppressing redundant noise in background regions.
\begin{itemize}
	\item $\mathit{Mask} \in \{0,1\}^{H \times W}$: Binary spatial filter generated by gradient thresholding.
	\item $\tau_d = 10000^{2d/D}$ ($d \in \{0,1,\dots,D/2\!-\!1\}$)
	\item $\odot$: Hadamard product for element-wise masking.
\end{itemize}

Meanwhile, heat regions (good embeddings) preserve multi-scale sinusoidal encoding characteristics, constructing target-sensitive context modeling. 

\subsubsection{HIDQ: HeatMap Induced High-Quality Queries for Decoder}
\label{subsubsec:HIDQ}

\textbf{HIDQ (Decoder-oriented)}: The HIDQ module transforms embedding features of the \textbf{mixed heatmap} ($H_{\text{mixed}}$) into \textbf{high-quality query} vectors.

The mixed heatmap undergoes linear transformation to generate initial queries:
\begin{equation}
	Q_{\text{dec}} = W'_Q E_{\text{mixed}},
\end{equation}
followed by a deformable attention mechanism:
\begin{equation}
	\mathit{DeformAttn}(Q_{\text{dec}}, K_{\text{enc}}, V_{\text{enc}}).
\end{equation}

HMPE-generated high-qulity queries (serving as the exclusive QUERY) suppress background-contaminated low-quality queries through attention reweighting, thereby simultaneously improving the decoder's precision in bounding box regression and robustness in semantic category prediction.

This work proposes that HMPE-generated high-quality queries significantly enhance object detection performance. 

In encoding, the MOHFE module innovatively integrates  class and bbox heatmaps, employs a gradient-based Mask filter to build robust positional encodings, and suppresses background noise while expanding receptive fields through feature disentanglement andmulti-scale embeddings. 

In decoding, the devised HIDQ converts mixed heatmap features into high-quality initial queries, leveraging deformable attention mechanisms to consolidate semantic-locational correlations, thereby substantially improving the detection head's regression accuracy for geometric attributes and semantic features.

\subsection{Linear-Snake Conv feature engineering}
\label{subsec:Linear-Snake}
\begin{figure}[H] 
	\centering
	\includegraphics[width=1.03\columnwidth]{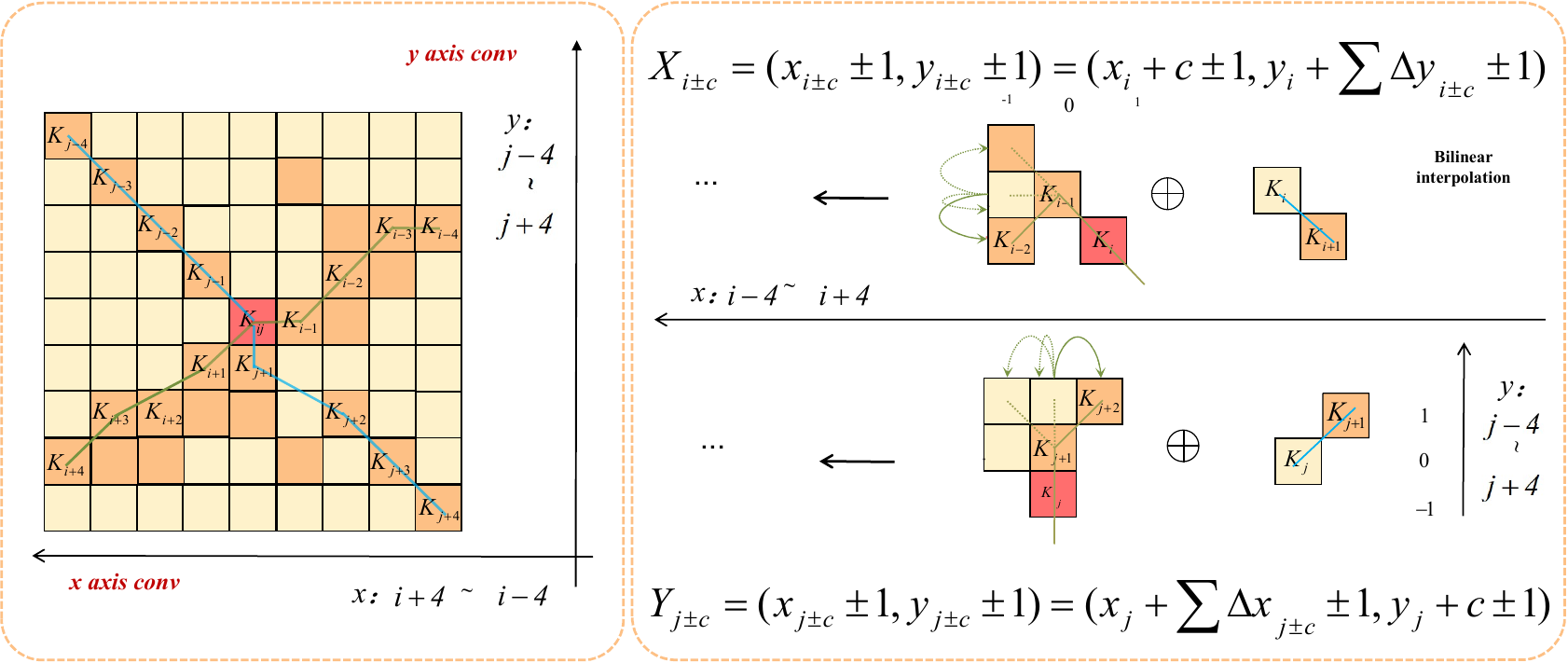}
	\caption{Dual-path complementary structure in a high dim structure.The \textcolor{green}{green} lines represent the feature extraction path for the snake section, while the \textcolor{blue}{blue} lines denote the path for the linear section.The red square denotes the center of convolution. The left image illustrates the efficient extraction of complex features under the dual-path complementary structure, and the right image simulates the extraction processes along the x-axis and y-axis,respectively.}
	\label{fig:axis-crop}
\end{figure}

\begin{figure}[t] 
	\centering
	\includegraphics[width=1\columnwidth]{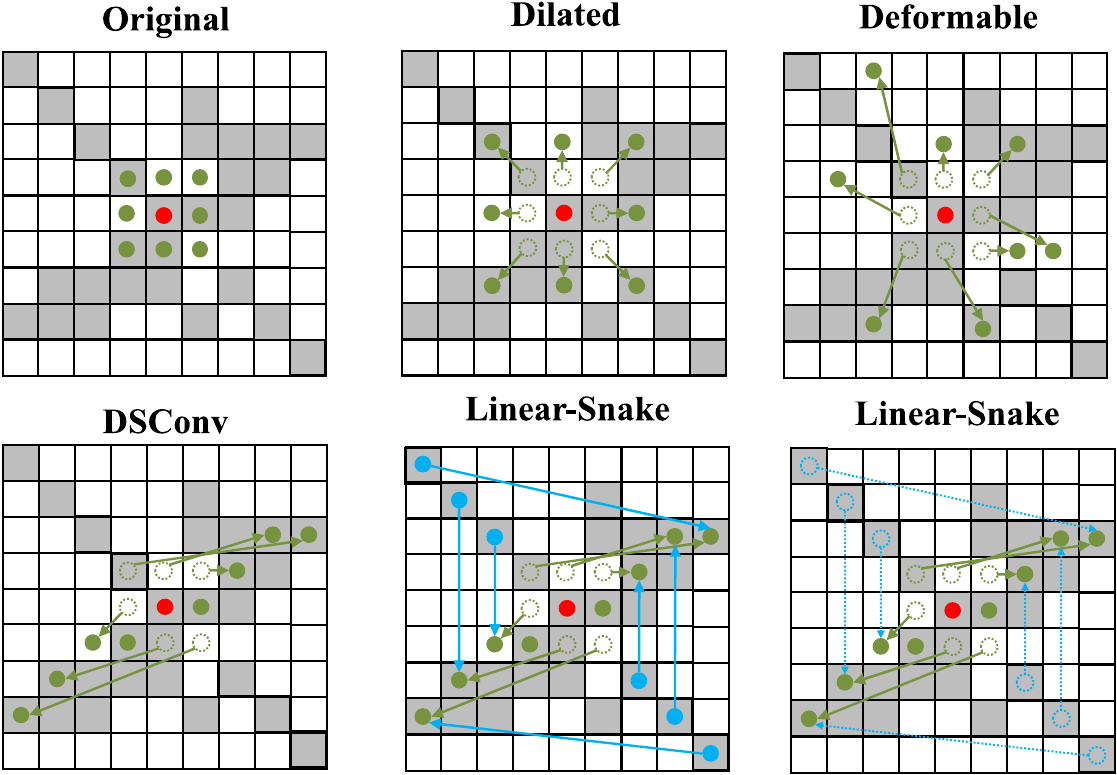}
	\caption{This demonstrates traditional convolution, dilated convolution, deformable convolution\cite{zhu2019deformable}, DSC convolution\cite{qi2023dynamic}, and two different variants of LSConv operating on a Linear-Snake over a 9x9 grid.\textcolor{blue}{Blue} represents the simulated path of Lsconv convolution, while \textcolor{green}{green} indicates the simulated path of other convolutions.}
	\label{fig:conv-crop}
\end{figure}
To address the challenges posed by extreme feature sparsity in complex small targets and general detection tasks, we propose LSConv (Linear-Snake Convolution), a novel feature engineering solution designed to capture geometrically intricate and sparse patterns. This module integrates into an encoder-decoder architecture through heatmap embedding, overcoming limitations of conventional convolution, Deformable Conv, Dilated Conv, and DSConv in balancing continuous curvilinear features and linear-semantic consistency.
 
Inspired by deformable receptive fields and dilated sampling strategies, LSConv introduces trainable deformation offsets ($\Delta$) predicted by a convolutional trainned predictor network, constrained by continuity regularization to avoid excessive deformation-induced receptive field expansion. For a 3×3 kernel operating within a 9×9 grid (x-axis example), each grid position is defined as $ {\Large X}_i \pm c = (x_i \pm c, y_i \pm c) $, where $ c \in {0,1,2,3,4} $ denotes the horizontal distance from the central grid. This cumulative offset mechanism restricts displacement within 9 standard convolution strides, preventing feature misalignment while preserving geometric precision. As figure \ref{fig:axis-crop}, LSConv employs a dual-path complementary architecture to simultaneously extract sparse curvilinear and continuous linear features:
\begin{itemize}
	
    \item A Dynamic Snake Convolution branch traces irregular curves through serpentine deformation.
	\item A Constrained Linear Convolution (Linear Conv) branch enforces linear-semantic consistency via fixed receptive fields. 
	
\end{itemize}
 	\begin{table*}[htbp]
	\centering
	\caption{Comparison of different methods on VOC07+12 And NWPU VHR-10 datasets.The optimal results are highlighted in red, and the suboptimal results are in blue.}
	\label{tab:comparison}
	\renewcommand{\arraystretch}{1.2} 
	\setlength{\tabcolsep}{3pt} 
	\begin{tabular}{l l c c c c c c c c c}
		\hline \hline
		\multirow{3}{*}{ } & \multirow{3}{*}{Method} & \multirow{3}{*}{Date} & \multirow{3}{*}{Backbone} & \multicolumn{4}{c}{Datasets} & \multirow{3}{*}{Epoch} & \multirow{3}{*}{GFLOPs} & \multirow{3}{*}{\begin{tabular}{c} Parameters \\ (MB) \end{tabular}} \\
		\cline{5-8}
		& & & & \multicolumn{2}{c}{PASCAL VOC} & \multicolumn{2}{c}{NWPU} & & & \\
		\cline{5-8}
		& & & & mAP & mAP@0.95 & mAP & mAP@0.95 & & & \\
		\hline
		\multirow{14}{*}{CNN} & SSD & 2016ECCV & RestNet18 & 51.01 & 22.98 & 40.34 & 16.01 & 75 & 353 & 26.29 \\
		& FCOS & 2019ICCV & ResNeXt & 67.41 & 43.64 & 84.35 & 55.68 & 75 & 32 & 31.8 \\
		& Retinanet & 2018CVPR & ResNet34 & 58.21 & 44.50 & 87.78 & 57.89 &75 & 39 & 36.1 \\
		& Faster R-CNN & 2017ICCV & VGG16 & 65.97 & 38.63 & 89.15 & 61.23 &75 & 63 & 41.1 \\
		& CenterNet & 2019ICCV & ResNet34 & 57.64 & 30.90 & 81.28 & 48.76 & 75 & 130 & 41.7 \\
		& MobilenetV3 & 2019ICCV & MBNV3 & 52.40 & 30.12 & 62.97 & 40.56 & 75 & 14 & 8.86 \\
		& YOLOv3 & 2018CVPR & ResNet50 & 64.08 & 43.76 & 84.50 & 52.34 & 75 & 112 & 43.1 \\
		& YOLOv4 & 2020CVPR & Darknet-53 & 62.70 & 49.90 & 87.80 & 60.89 & 75 & 101 & 44.9 \\
		& YOLOv5 & 2020ArXiv & CSPDarknet53 & 67.10 & 49.73 & 89.11 & 65.43 & 75 & 109 & 46 \\
		& YOLOv6 & 2022ArXiv & CSPDarknet53 & 67.16 & 54.28 & 89.78 & 52.10 &75 & 150 & 59 \\
		& YOLOv7 & 2023CVPR & CSPRep53 & 67.80 & 44.30 & 91.03 & 63.12 &75 & 104 & 36 \\
		\multirow{10}{*}{Transformer} & YOLOv8 & 2023Arxiv & CSPDarknet18 & 53.01 & 36.66 & 90.43 & 60.80 & 75 & 32 & 16.1 \\
		& YOLOv11 & 2025ArXiv & CSPRep53 & \textcolor{red}{71.70} & 47.3 & \textcolor{blue}{92.68} &\textcolor{red}{ 68.89} & 75 & 105 & 38.8 \\
		\hline
		
		& DETR & 2020ECCV & ResNet50 & 62.40 & 42.00 & 81.28 & 35.9 & 125 & 101 & 36.74 \\
		& Deformable DETR & 2021ICLR & ResNet50 & 60.56 & \textcolor{red}{51.92} & 79.30 & 40.50 &125 & 179 & 39.83 \\
		& PR-Deformable DETR* & 2024GRSL & ResNet50 & / & / & 88.30 & 43.20 & 125 & 151 & 46.07 \\
		& RT-Detr & 2024CVPR & HGNetv1 & 69.41 & 50.59 & 92.60 & 60.26 & 100 & 136 & 163 \\
		\rowcolor{LightGreen} 
		& Our work & 2025 & HGNetv2 & \textcolor{blue}{70.61} & \textcolor{blue}{51.50} & \textcolor{red}{94.50} & \textcolor{blue}{67.20} & 100 & 57 & 77.3 \\
		\hline \hline
	\end{tabular}
\end{table*}

 LSConv uses bilinear interpolation to sample deformed features— horizontal (x-axis) processing employs a 3×1 strip kernel to aggregate continuous pixel data with locked vertical offsets. According to Equation (x), this process can be further detailed as follows:

\hspace{-2em}
\vspace{-1em}
\begin{align}
	X_ {i\pm c} = \begin{cases}
		\begin{aligned}[t]
			&(x_{i + c} + 1, y_{i + c} + 1) \\
			&\quad = \left(x_i + c + 1, y_i + \sum \Delta y_{i + c} + 1\right),
		\end{aligned} \\[2ex] 
		\begin{aligned}[t]
			&(x_{i - c} - 1, y_{i - c} - 1) \\
			&\quad = \left(x_i + c - 1, y_i + \sum \Delta y_{i - c} - 1\right).
		\end{aligned}     
	\end{cases}
\end{align}

While vertical (y-axis) processing utilizes a $1 \times 3$ kernel under analogous constraints, the similar formula can be written as:
\hspace{-2em}  
\begin{equation}
	Y_{j \pm c} = \begin{cases}
		\begin{aligned}[t]
			&(x_{j + c} + 1, y_{j + c} + 1) \\
			&\quad = \left(x_j + \sum \Delta x_{j + c} + 1, y_j + c + 1\right),
		\end{aligned} \\ 
		\begin{aligned}[t]
			&(x_{j - c} - 1, y_{j - c} - 1) \\
			&\quad = \left(x_j + \sum \Delta x_{j - c} - 1, y_j + c - 1\right).
		\end{aligned}     
	\end{cases}
\end{equation}

  Weighted fusion of snake-path offsets (adaptive to dense curvilinear regions) and linear-path offsets (biased toward straight features) ensures synergistic feature extraction. The snake branch dominates in curved regions through staggered offset accumulation, while the linear branch prioritizes straight-line continuity, optimizing geometry-aware alignment. This dual-mode mechanism balances geometric specificity (for small-target semantics) and structural coherence (for linear semantics) via coordinated path interaction, achieving superior performance in complex detection scenarios by adaptively refining receptive field coverage while mitigating feature dispersion and misregistration.

	\begin{table*}[h]
		\centering
		\caption{NWPU VHR-10 Ablation Experiments}
		\begin{tabular}{ccccc|cc|ccc} 
			\toprule
			\toprule			
			\multirow{2}{*}{Model} & \multirow{2}{*}{DSconv} & \multirow{2}{*}{Linear-Snake} & \multicolumn{2}{c|}{HeatMap Embedding} & \multirow{2}{*}{Precision(\%)} & \multirow{2}{*}{Recall(\%)} & \multirow{2}{*}{mAP(\%)} & \multirow{2}{*}{mAP@95(\%)} \\
			\cmidrule(lr){4-5} 
			& & & MOHFE & HIDQ & & & & \\
			\midrule
			Baseline & - & - & - & - & 88.9 & 89.9 & 92.6 & 60.2 \\
			\midrule
			& $\checkmark$ & - & - & - & 89.0$\uparrow$+0.10 & 90.10$\uparrow$+0.20 & 92.71$\uparrow$+0.11 & 62.32$\uparrow$+2.12 \\
			& $\checkmark$ & $\checkmark$ & - & - & 90.74$\uparrow$+1.84 & 90.62$\uparrow$+0.72 & 93.91$\uparrow$+1.31 & 63.69$\uparrow$+3.49 \\
			Baseline+ Parts & $\checkmark$ & $\checkmark$ & $\checkmark$ & - & 90.75$\uparrow$+1.85 & 90.51$\uparrow$+0.61 & 94.18$\uparrow$+1.58 & 63.94$\uparrow$+3.74 \\
			\rowcolor{LightGreen} 
			& $\checkmark$ & $\checkmark$ & $\checkmark$ & $\checkmark$ & 93.17$\uparrow$+4.97 & 90.40$\uparrow$+0.50 & 94.50$\uparrow$+1.90 & 67.20$\uparrow$+6.90 \\ 
			\bottomrule
			\bottomrule
		\end{tabular}
	\end{table*}

	\begin{table*}[h]
		\centering
		\caption{Voc07+12 Ablation Experiments}
		\begin{tabular}{ccccc|cc|ccc} 
			\toprule
			\toprule
			\multirow{2}{*}{Model} & \multirow{2}{*}{DSconv} & \multirow{2}{*}{Linear-Snake} & \multicolumn{2}{c|}{HeatMap Embedding} & \multirow{2}{*}{Precision(\%)} & \multirow{2}{*}{Recall(\%)} & \multirow{2}{*}{mAP(\%)} & \multirow{2}{*}{mAP@95(\%)} \\
			\cmidrule(lr){4-5} 
			& & & MOHFE & HIDQ & & & & \\
			\midrule
			Baseline & - & - & - & - & 76.7 & 62.9 & 69.4 & 50.5 \\
			\midrule
			& $\checkmark$ & - & - & - & 77.20$\uparrow$+0.50 & 62.93$\uparrow$+0.03 & 70.29$\uparrow$+0.89 & 51.17$\uparrow$+0.67 \\
			& $\checkmark$ & $\checkmark$ & - & - & 78.41$\uparrow$+1.71 & 63.10$\uparrow$+0.20 & 70.19$\uparrow$+0.79 & 51.41$\uparrow$+0.91 \\
			Baseline+ Parts & $\checkmark$ & $\checkmark$ & $\checkmark$ & - & 78.52$\uparrow$+1.82 & 63.55$\uparrow$+0.65 & 70.47$\uparrow$+1.07 & 51.33$\uparrow$+0.83 \\
			\rowcolor{LightGreen} 
			& $\checkmark$ & $\checkmark$ & $\checkmark$ & $\checkmark$ & 78.80$\uparrow$+2.10 & 63.72$\uparrow$+0.82 & 70.61$\uparrow$+1.21 & 51.50$\uparrow$+1.00 \\
			\bottomrule
			\bottomrule
		\end{tabular}
	\end{table*}
	
	\label{vis1}
	\begin{figure*}[ht] 
		\centering
		\includegraphics[width=1\textwidth]{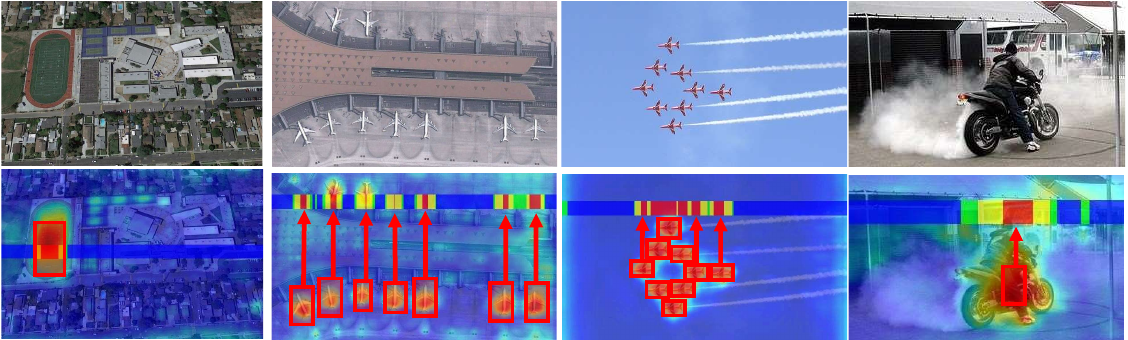}
		\caption{Visualization of HeatMap Embedding.It includes a comparison between complex scenarios and general detection.}
		\label{fig:vis2}
	\end{figure*}

\section{EXPERIMENTAL RESULTS AND ANALYSIS}

\subsection{Datasets and Evaluation Criteria}
Our evaluation is conducted on two publicly available benchmarks: one is a small object detection benchmark, NWPU VHR-10~\cite{cheng2016learning}, and the other is a general detection dataset, Pascal VOC~\cite{everingham2010pascal}.

The Pascal VOC dataset encompasses various versions from each year's competition, with the benchmarks being the 2007 and 2012 editions. We have merged VOC07 and VOC12, which consist of 20 categories. The training data includes 16,551 images with a total of 40,058 objects, and the test data comprises 4,952 images containing 12,032 objects. The NWPU VHR-10 dataset is geared towards aerial and remote sensing detection, featuring 650 images containing objects and 150 background images, totaling 800 images. These were cropped from Google Earth and the Vaihingen dataset and then manually annotated by experts. The remaining detection standards strictly follow the Paper\cite{zhao2024detrs}.
\subsection{Comparison with State-of-the-art Methods}
To verify the effectiveness of the proposed method in both general and small detection datasets, we compare it with numerous existing object detection methods, including both CNN-based and Transformer-based methods. Among the CNN-based methods are: FCOS\cite{tian2019fcos}, RetinaNet\cite{lin2017focal}, Faster R-CNN\cite{ren2015faster}, CenterNet\cite{duan2019centernet}, MobileNet-V3\cite{howard2019searching}, YOLOv3\cite{redmon2018yolov3}, YOLOv4\cite{bochkovskiy2020yolov4}, YOLOv5\cite{benjumea2021yolo}, YOLOv6\cite{li2022yolov6}, YOLOv7\cite{wang2023yolov7}, YOLOv8, YOLOv11\cite{khanam2024yolov11},Transformer-based methods  are DETR, Deformable DETR, PR-Deformable DETR*, RT-Detr, and Our Work.The hyperparameters for the aforementioned CNN-based tasks are identical, while the Transformer-based hyperparameters align with Our work.

Table ~\ref{tab:comparison} presents the experimental results of using different object detection methods on Two datasets, where CNN-based models typically converge within 75 epochs during testing, whereas transformer-based models require training for 100 to 125 epochs to achieve convergence.
Transformer-based detectors exhibit data-hungry characteristics, requiring extensive training support. For instance, their lower mAP95 on NWPU VHR-10 (35.9\% for DETR and 67.20\% for our work) highlights limited adaptability to small objects under sparse data.
HMPE Advantage: The proposed method achieves 94.50\%mAP and 67.20 \%mAP95 on NWPU (outperforming YOLOv11's 92.68\%/68.89\%) , demonstrating that HMPE effectively injects high-quality queries into encoder-decoder architectures by Leveraging conditional coupling between heatmaps and positional encodings to stabilize query generation with minimal training samples (evidenced by +3.3\% mAP95 over YOLOv7) , even outperforming computationally heavy models like RT-Detr's GFLOPs=136G more than 57G.Notably, the proposed method achieves 58.8M parameter reduction (77.3M vs. RT-Detr’s 163M) while maintaining superior NWPU mAP95 (+6.94\% over RT-Detr).

\subsection{Ablation Studies }

\textbf{Results on NWPU VHR-10}.
The systematic experiments validate the progressive enhancement mechanism of each component on the NWPU VHR-10 dataset. The baseline model achieves 92.6\% mAP and 60.2\% mAP95 without any proposed modules. The incorporation of Linear-Snake Conv significantly improves fine-grained geometric perception through dual-path complementary architecture feature extraction, boosting mAP95 by 3.4\% to 63.6\% alongside a 1.8\% precision improvement to 90.7\%. Activating MOHFE elevates mAP to 94.1\% through heatmap-driven positional encoding optimization, representing a 1.5\% gain over baseline. Finally, integrating the HIDQ module achieves comprehensive performance breakthroughs: the heatmap-induced query refinement sharpens detection precision to 93.1\% (+4.97\% absolute improvement) while propelling mAP95 to 67.2\% (+6.94\% over baseline). This progressive improvement demonstrates that MOHFE enhances background suppression through semantics-sensitive positional encoding, while HIDQ further optimizes object localization precision via decoder-side heatmap-guided high  qulity queries refinement.

\noindent\textbf{Results on PASCAL VOC}.
The progressive implementation demonstrates consistent improvements across all metrics. Starting from the baseline (69.4\% mAP, 50.5\% mAP95), the sequential integration of  Linear-Snake (+0.91\% mAP95) confirms their geometric modeling benefits. While MOHFE strengthens positional-semantic coupling (+0.83\% mAP95), the complete framework with HIDQ achieves optimal performance: +2.10\% precision (78.80\%) and +1.21\% mAP (70.61\%) with 51.52\% mAP95 (+1.02\%). Notably, the recall-precision equilibrium improves progressively, maintaining a 0.82\% recall gain. Though the gains are relatively smaller than NWPU due to the dataset's lower object density, these results validate the framework's cross-domain robustness, particularly the critical role of HIDQ in distilling heatmap-derived queries.

Systematic experiments on the NWPU VHR-10 and PASCAL VOC datasets validate the progressive enhancement mechanisms of each component. HIDQ significantly improves the quality of query representations in both datasets, particularly achieving substantial performance gains in the high-density target scenarios of NWPU. MOHFE enhances background suppression through optimized positional encoding driven by semantic detection. Linear-Snake Conv improves feature extraction for small objects through dual-path complementary architecture feature extraction, regardless of whether the scenario involves dense or sparse targets.

\subsection{Visualization of HeatMap Embedding}

As shown in Figure \ref{fig:vis2} ,two images on the left represent the standard visualization of NWPU VHR-10, while the two on the right depict the standard visualization of PASCAL VOC2012. 
The top four images represent the original input to the detector, whereas the bottom four images illustrate the visualization results that incorporate the heatbar after upscaling the encoder by a factor of six to match the dimensions of the original image.
Consistent with the description in \ref{fig:vision of HMPE}, the embeddings in the heatbar exhibit a pattern of being "hot" in the middle and "cold" at both ends, indicating their efficient embedding into the Query.
The presence of padding during downsampling, along with resizing, may cause the heatmap visualization of the embeddings to potentially exceed mapping boundaries, leading to possible discrepancies between the visualized and actual embedded positions.

\subsection{Ablation Study on MultiHead Of Detr}
\begin{table}[ht]
	\setlength{\aboverulesep}{0.5ex}   
	\setlength{\belowrulesep}{0.5ex}   
	\caption{MultiHead Layer ablation in decoder. Det$^k$ denotes detectors with k decoder layers.  Det$^{3\sim7}$ indicates that the output results from the 3rd to the 7th layers of the detection head have reached the upper limit across a total of 5 layers.Red represents the upper limit of the most efficient layers. }
	\renewcommand{\arraystretch}{1.2}
	\label{tab:params} 
	\centering
	\begin{tabular}{c|c c c c|c c c c}
		\toprule
		\toprule 
		\multirow{2}{*}{Epoch} & \multicolumn{4}{c}{AP(\%)} & \multicolumn{4}{c}{GFLOPs} \\
		\cline{2-5} \cline{6-9}
		& Det$^0$ & Det$^1$ & Det$^2$ & Det$^{3\sim7}$ & 0 & 1 & 2 & 3$\sim$7 \\
		\midrule
		
		100 & 66.7 & 69.9 & \textcolor{red}{70.5} & 70.5 &\multicolumn{4}{c}{ } \\
		50 & 62.4 & 65.6 & 63.1 & \textcolor{red}{66.8} & 53.7 & 55.4 & 56.2 & 57.0 \\
		25 & 40.7 & 41.3 & 43.2 & \textcolor{red}{47.9} &\multicolumn{4}{c}{ } \\
		Fewshot & 38.9 & 42.9 & 43.8 & \textcolor{red}{43.9} &\multicolumn{4}{c}{ } \\
		\bottomrule
		\bottomrule
	\end{tabular}
\end{table}


\noindent\textbf{Results on Decoder MultiHead Layer}.
 Epoch indicates the number of training iterations for decoder layers.All results are obtained under HMPE-driven high-quality query augmentation. The ablation study demonstrates that utilizing only 3 decoder heads ( Det$^{3\sim7}$) with HMPE-driven query optimization achieves state-efficient detection in few-shot scenarios (43.9\% AP), while maintaining full-supervision performance parity (70.5\% AP vs. 69.9\% AP for 8 heads). As shown in Table~\ref{tab:params} ,this configuration reduces computational costs by 1.4\% (57.0→56.2 GFLOPs) and eliminates redundant parameters, proving that HMPE enables lightweight decoders to concentrate computational resources on geometrically-critical regions without multi-head redundancy.For detection tasks, few-shot learning requires no more than three decoder heads, while increasing training epochs enables performance saturation with merely two heads.
 
 \noindent\textbf{Results on Encoder-Decoder architecture params}.
 The encoder, designed for feature extraction, uses 3.01 MB, while the decoder, handling query optimization and detection, requires 16.8 MB. This allocation highlights the decoder's critical role in precise localization and classification, ensuring efficient performance with a balanced parameter count.

\section{CONCLUSION AND OUTLOOK}
We innovatively designed heatmap embedding and its visualization, enabling an observable insight into the embedded information, and utilized it to fine-tune the parameters. We proposed Multi-Scale ObjectBox-Heatmap Fusion Encoder (MOHFE) and HeatMap Induced High-Quality Decoder Queries (HIDQ) modules optimize the embedding for the encoder and decoder, respectively, effectively improving query quality and reducing redundant computations. Additionally, the LSConv module further enhances the feature extraction capability for complex small objects, providing a more efficient solution for small object detection tasks. Experimental results demonstrate significant performance improvements on both the small object dataset NWPU VHR-10 and the general dataset PASCAL VOC, along with a notable reduction in computational costs for both inference and training. These innovations not only provide a new technical pathway for the field of small object detection but also offer valuable insights for other vision tasks. In the future, we will continue to explore the potential applications of heatmap embedding in other visual tasks and further optimize the model architecture to address more complex real-world scenarios.


\newpage
\newpage  
\bibliographystyle{ACM-Reference-Format}
\bibliography{refs}

\end{document}